\documentclass[a4paper]{article}

\usepackage{INTERSPEECH2022}
\usepackage{multirow}
\newcommand{\comment}[1]{}
\usepackage{arydshln}
\usepackage{booktabs,arydshln}
\usepackage{graphicx}
\makeatletter
\def\adl@drawiv#1#2#3{%
        \hskip.5\tabcolsep
        \xleaders#3{#2.5\@tempdimb #1{1}#2.5\@tempdimb}%
                #2\z@ plus1fil minus1fil\relax
        \hskip.5\tabcolsep}
\newcommand{\cdashlinelr}[1]{%
  \noalign{\vskip\aboverulesep
           \global\let\@dashdrawstore\adl@draw
           \global\let\adl@draw\adl@drawiv}
  \cdashline{#1}
  \noalign{\global\let\adl@draw\@dashdrawstore
           \vskip\belowrulesep}}
\makeatother

\title{Zero-Shot Cross-lingual Aphasia Detection using \\ Automatic Speech Recognition}
\name{Gerasimos Chatzoudis$^1$, Manos Plitsis$^{1,3}$, Spyridoula Stamouli$^1$, Athanasia–Lida Dimou$^1$, Athanasios Katsamanis$^{1,2}$, Vassilis Katsouros$^1$}
\address{
  $^1$Institute for Language and Speech Processing, Athena Research Center, Athens, Greece\\
  $^2$ Behavioral Signal Technologies, Los Angeles, CA, USA\\
  $^3$ National and Kapodistrian University of Athens, Athens, Greece}
\email{\{gerasimos.chatzoudis, manos.plitsis, pstam, ndimou, nkatsam, vsk\}@athenarc.gr}

\begin{document}

\maketitle
\begin{abstract}
  Aphasia is a common speech and language disorder, typically caused by a brain injury or a stroke, that affects millions of people worldwide. Detecting and assessing Aphasia in patients is a difficult, time-consuming process, and numerous attempts to automate it have been made, the most successful using machine learning models trained on aphasic speech data. Like in many medical applications, aphasic speech data is scarce and the problem is exacerbated in so-called ``low resource" languages, which are, for this task, most languages excluding English. We attempt to leverage available data in English and achieve zero-shot aphasia detection in low-resource languages such as Greek and French, by using language-agnostic linguistic features. Current cross-lingual aphasia detection approaches rely on manually extracted transcripts. We propose an end-to-end pipeline using pre-trained Automatic Speech Recognition (ASR) models that share cross-lingual speech representations and are fine-tuned for our desired low-resource languages. To further boost our ASR model’s performance, we also combine it with a language model. We show that our ASR-based end-to-end pipeline offers comparable results to previous setups using human-annotated transcripts.

\end{abstract}
\noindent\textbf{Index Terms}: Pathological speech assessment, aphasia, Greek, AphasiaBank, Disordered speech recognition, machine learning, zero-shot classification

\section{Introduction}
Aphasia is a common speech and language disorder resulting from a brain injury that is usually caused by a stroke. According to the National Aphasia Association, at least 2,000,000 people in the USA and 250,000 in the UK are currently being affected by aphasia and this number is increasing at a rate of 180,000 people who acquire aphasia each year  \cite{national_aphasia_association}. People with Aphasia (PWA) face communication difficulties making social interaction inefficient due to their impaired speech production and/or language understanding. To mitigate these communication inefficiencies, PWA need to attend long-term, intensive targeted therapies with expertly trained Speech-Language Pathologists (SLPs) \cite{brady2016speech}. Such conventional in-clinic therapies are time-consuming, expensive, and, in some circumstances, unattainable for a lot of people due to the lack of local and long-term options.

\begin{figure}[t]
  \centering
  \includegraphics[width=\linewidth]{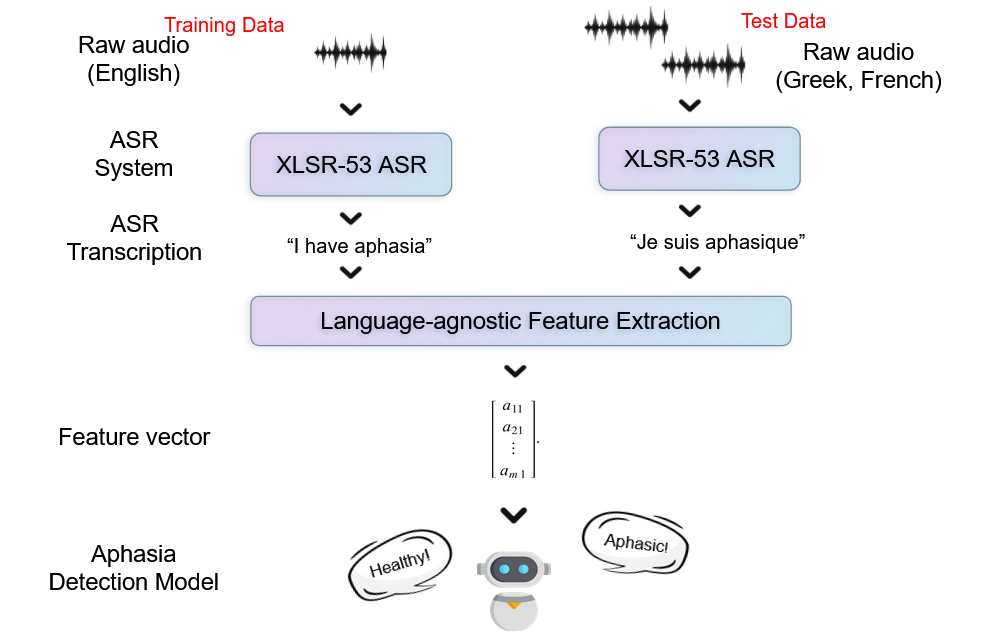}
  \caption{Schematic diagram of the proposed end-to-end pipeline using language-agnostic linguistic features. Our classifier is trained using English data (left) and evaluated in our low-resource languages, i.e. French and Greek (right).}
  \label{fig:speech_production}
\end{figure}

Towards this end, several attempts have been made to detect aphasia from spontaneous speech \cite{qin_automatic_2018}, and develop end-to-end automatic assessment systems \cite{le_automatic_2018, qin_automatic_2020}. These studies indicate that it is possible to automatically detect and evaluate aphasia using acoustic and linguistic features from either manual or ASR transcriptions. However, these works only focus on languages with sufficient data, such as English. Data scarcity of aphasic speech in low-resource languages makes it difficult to democratize these Machine Learning approaches, and enable aphasia assessment for everyone. 
One solution to the problem of data scarcity is given by cross-lingual methods that leverage language-agnostic features from resource-rich languages, and transfer them to low-resource settings. Previous attempts have been made 
to achieve few-shot and zero-shot aphasia detection from English to both closely related (e.g., French) and more distant languages (e.g., Mandarin) \cite{pmlr-v116-balagopalan20a, shivkumar_blabla_2020}. However, these studies rely heavily on manual, human-annotated transcriptions by professional SLPs, a procedure that is costly and tedious, even for a small amount of data. 

For that purpose, we investigate extending current cross-lingual aphasia detection pipelines by incorporating Automatic Speech Recognition models for two closely related to English low-resource languages, i.e., French and Greek. Since training an ASR domain-specific model for these languages requires a considerable amount of collected speech data that is currently unavailable, we use large pre-trained ASR models that have demonstrated significant capabilities across several domains \cite{xiao_scaling_2021}. Specifically, we leverage pre-trained XLSR-53 architectures \cite{DBLP:journals/corr/abs-2006-13979} based on the Wav2Vec2.0 audio representation \cite{NEURIPS2020_92d1e1eb}, that have been proven to work for other low-resource languages on AphasiaBank, i.e., Spanish \cite{torre_improving_2021}. To augment ASR performance, we incorporate language models trained on out-of-domain data for each language in our architecture.

Our main contribution is an end-to-end cross-lingual aphasia detection system for Greek and French, using language-agnostic linguistic features directly extracted from pre-trained ASR systems (as shown in Figure \ref{fig:speech_production})\footnote{The source code for reproducing all our experiments can be found at https://gitlab.com/ilsp-spmd-all/public/crosslingual-aphasia-detection}. This research activity is part of the ongoing project “PLan-V: A Speech and Language Therapy Platform with Virtual Agent”, which aims to develop a technologically-assisted speech and language intervention platform for people with chronic neurogenic communication disorders, such as aphasia, integrating a system for the automated assessment of aphasia severity\footnote{This research has been co-financed by the European Regional Development Fund of the European Union and Greek national funds through the Operational Program Competitiveness, Entrepreneurship and Innovation, under the call RESEARCH—CREATE—INNOVATE (Project title: PLan-V: A Speech and Language Therapy Platform with Virtual Agent, Project code: T2EDK-02159).}. 

\section{Related Work}
\label{related}
The first attempts at aphasic speech recognition involved short, prompted speech segments \cite{abad_automatic_2013}.
    One of the first papers to combine aphasic speech recognition with aphasia severity assessment was \cite{le_automatic_2018}. In this work, the researchers use human-annotated and ASR-based transcriptions from the English AphasiaBank, to extract certain linguistic features and develop a model that ultimately predicts a speaker's AQ score. A similar ASR-based approach, using features extracted with deep neural networks, was adopted by \cite{qin_automatic_2020} for the Cantonese AphasiaBank. Both teams focused more on the robustness of the features used for the predicting model in their respective language, rather than on speech recognition performance. Both teams have also attempted to use out-of-domain speaker data to improve ASR performance in various ways \cite{le_improving_2016,qin_automatic_2018-1}.
    Recently, FacebookAI researchers have improved ASR performance in the English AphasiaBank by using neural models trained with vast amounts of data, and then adapted to various domains \cite{xiao_scaling_2021}. Semi-supervised learning was also used recently in an attempt to improve ASR performance in both English and Spanish \cite{torre_improving_2021}.

    Several attempts have also been made to classify different aphasia types from manual transcripts. Researchers in \cite{fraser_automated_2014} and  \cite{fromm_enhancing_2021} use similar features extracted from the AphasiaBank manual transcriptions. Interestingly, in \cite{fromm_enhancing_2021} the two features that were found most significant were the number of words and number of closed terms in the Cinderella story, which are respective measures of productivity and grammaticality, qualities that we also find important when assessing aphasic status, while in \cite{fraser_automated_2014} we see similar patterns in the most important features.
    
    Our work attempts to solve the problem of training an aphasia detection model from speech, for a low-resource language, for which there is a very limited amount of data available (or even no data at all). We are currently aware of two works that have attempted to solve this problem, both, however, using linguistic features extracted only from human-annotated textual transcriptions of available speech data \cite{pmlr-v116-balagopalan20a, shivkumar_blabla_2020}. Both use transfer learning techniques to apply zero-shot and few-shot aphasia classification. Balagopalan et al. in \cite{pmlr-v116-balagopalan20a} claim that it may be possible to transfer features from languages that are not distant and this is also validated in \cite{shivkumar_blabla_2020}, where a wider set of linguistic features is also used. Last, in \cite{themistocleous_automatic_2021}, the authors used cross-lingual linguistic features to assess morphosyntactic production in agrammatic aphasia.
    

\section{Datasets}
Our dataset consists of speech samples from the AphasiaBank database for English and French, as well as a Greek corpus collected by various researchers. The amount of data available for each language can be seen in Table \ref{table:dataset}.

AphasiaBank is a shared database of multimedia interactions with the purpose of studying aphasia communication \cite{macwhinney_aphasiabank_2011,forbes_aphasiabank_2012}. It consists of a collection of transcribed audio and/or video recordings of interviews between clinicians and aphasic or healthy subjects. Depending on the discourse language and the contributing research group, various speech elicitation protocols were followed,  with the majority of protocols being variants of the AphasiaBank data collection protocol. In particular, we used speech samples from the English and French AphasiaBank subgroup, and specifically, only data that were collected according to the AphasiaBank protocol. AphasiaBank's transcriptions are produced using the CHAT transcription format \cite{macwhinney2000childes} that includes, besides textual transcriptions, a variety of symbols denoting non-speech sounds such as laughs and coughs, as well as a system for transcribing paraphasias, or mispronounced words, in the International Phonetic Alphabet.

Our Greek data come from two main sources: a) the GREECAD corpus \cite{varlokosta2016greek}, a dataset of spoken narratives collected from 22 Greek-speaking PWA and 10 unimpaired controls using a protocol of 4 narrative tasks and b) the PLan-V project corpus \cite{stamouli2022planv}, a dataset of spoken narratives collected from 10 unimpaired controls using an augmented narrative discourse elicitation protocol, which includes 4 tasks from the AphasiaBank protocol and 3 tasks from the GREECAD protocol. In both datasets, spoken language samples have been manually transcribed in an orthographic format and segmented into time-aligned utterances, following the CHAT guidelines for utterance segmentation. We chose not to incorporate the Greek datasets of the AphasiaBank repository, mainly because of their conversational form, in contrast to the GREECAD and PLan-V data, which represent different types of monologic narrative discourse, with minimal interventions from the investigator.


\begin{table}[th]
  \caption{ Dataset Statistics}
  \label{table:dataset}
  \centering
  \begin{tabular}{ l c c c}
  
    \toprule
    \multicolumn{1}{c}{\multirow{2}{*}{\textbf{Language}}} &
    \multicolumn{3}{c}{\textbf{Number of Speakers}}\\
    \cline{2-4} 
    & \multicolumn{1}{c}{\textbf{Total}} &
    \multicolumn{1}{c}{\textbf{Control}} &
    \multicolumn{1}{c}{\textbf{Aphasia}} \\
    \midrule
    English & 705 & 245 & 460 \\
    French  & 27 & 14 & 13 \\ 
    Greek   & 42 & 20 & 22 \\
    \bottomrule
  \end{tabular}
  
 \end{table}

\section{Methods}

\subsection{Preprocessing}
All available transcripts in each language contain various annotations, such as target phrases and pauses. These annotations, although useful, are unavailable in our automated assessment system, so we remove them from all transcripts using PyLangAcq \cite{lee-et-al-pylangacq:2016}. We only retain the cleaned transcripts, with their corresponding duration information.  We refer to these as oracle transcripts. For each oracle transcript, we downsample its corresponding audio to 16 kHz. Following data cleaning, we pass both oracle and ASR transcripts through a part-of-speech tagging, tokenization and lemmatization module. Tokens that correspond to out-of-vocabulary words are not omitted from the process. For English and French, we employ spaCy's transformer-based models\footnote{https://spacy.io/}, whereas for Greek, we use the Neural NLP toolkit for Greek \cite{10.1145/3411408.3411430}.

\subsection{Automatic Speech Recognition Models}
 Our suggested ASR model uses the XLSR-53 architecture \cite{DBLP:journals/corr/abs-2006-13979} which is based on the Wav2Vec2.0 audio representation \cite{NEURIPS2020_92d1e1eb} and was trained with 56,000 hours of speech data in 53 languages. Because it is based on a language-agnostic representation of speech, this architecture is particularly well-suited for multi-language tasks, where it can be fine-tuned for each language using very few data instances. Specifically, we used fine-tuned XLSR-53 models for each language, that were pre-trained on the Common Voice dataset \cite{commonvoice:2020}. Hugging Face provides open-source versions of these models in English \cite{grosman2021Wav2Vec2-large-xlsr-53-english}, French \cite{grosman2021Wav2Vec2-large-xlsr-53-french}, and Greek \cite{grosman2021Wav2Vec2-large-xlsr-53-greek}.

To further optimize the performance of the fine-tuned XLSR-53 models, we combined them with a language model for each language and employed a beam search decoding strategy. Each experiment used a beam width of 10. To train the language model, we first collected textual data from Wikipedia \cite{wikidump} and then trained Kneser-Ney smoothed n-gram language models using the KENLM toolkit \cite{10.5555/2132960.2132986}. Specifically, we trained a 3-gram model for all languages. The language models are integrated into the XLSR-53 architecture using the pyctcdecode library\footnote{https://github.com/kensho-technologies/pyctcdecode}, a fast and feature-rich CTC beam search decoder for speech recognition in Python. We evaluate our models on both Word-Error-Rate (WER) and Character-Error-Rate (CER) metrics \cite{wer-cer}.

\subsection{Language-agnostic Feature Extraction}
Previous works in automatic assessment of aphasia indicate that using linguistic features is considered standard practice (see Section \ref{related}). Given the condition's influence on the brain's language centers it is necessary for any reliable system to be able to access semantic and grammatical information. Over 500 linguistic indicators have been applied to assess spoken language ability and the efficacy of PWA interventions \cite{stamouli2022planv}.

Following recent works in the analysis of discourse production  \cite{stark2019comparison,fromm2021enhancing, stamouli2022planv}, we focus on linguistic features that evaluate 6 language abilities, i.e., linguistic productivity, content richness, fluency, syntactic complexity, lexical diversity, and gross output. A complete list of the extracted features is presented in Table \ref{table:features}. These make up a vector of size 24 per speaker. 

We chose to extract all features at the speaker level to minimize variability due to utterance length and number of stories between speakers. Our end-to-end approach can help minimizing the impact of this type of variability that is very common across datasets due to the different transcription guidelines that may have been followed in each case.

\comment{In quantitative cross-lingual aphasia assessment attempts, part-of-speech tagging related features \cite{pmlr-v116-balagopalan20a} and features such as noun-verb ratio, pronoun-noun ratio, Yngve depth, pronoun rate, and content density were selected \cite{shivkumar_blabla_2020}.}

\begin{table}[th]
  \caption{Language-agnostic linguistic features across 6 language ability categories.}
  \label{table:features}
  \centering
  \resizebox{\linewidth}{!}{\begin{tabular}{ l l }
    \toprule
    \multicolumn{1}{l}{\textbf{Language ability}} & 
    \multicolumn{1}{l}{\textbf{Feature}} \\
    \midrule
    Linguistic Productivity &    Mean Length of Utterance              \\
    \hline
    Content Richness &    Verb/Word Ratio \\
     &    Noun/Word Ratio \\
     &    Adjective/Word Ratio \\
     &    Adverb/Word Ratio \\
     &    Preposition/Word Ratio \\
     &    Propositional density \\
    \hline
    Fluency &    Words per Minute \\
    \hline
    Syntactic Complexity &    Verbs per Utterance \\
     &    Noun Verb Ratio  \\
     &  Open-closed class words\\
     &    Conjunction/Word Ratio \\
     &   Mean Clauses per utterance  \\
     &  Mean dependent clauses \\
     &  Mean independent clauses \\
     &  Dependent to all clauses ratio \\
     & Mean Tree height \\
     &  Max Tree depth \\
     &  Number of independent clauses \\
     &  Number of dependent clauses \\
     \hline
     
    Lexical Diversity   & Lemma/Token Ratio\\
     &    Words in Vocabulary per Words \\
    &    Unique words in vocabulary per Words \\
    \hline
    Gross output   &  Number of words \\

    \bottomrule
  \end{tabular}}
  
\end{table}      

\subsection{Cross-lingual aphasia detection}
We initially examine the feasibility of aphasia detection in English prior to developing our transfer learning pipeline across languages. We compare multiple classification models in this monolingual scenario, including SVM \cite{cortes1995support}, XGBoost \cite{Chen:2016:XST:2939672.2939785}, Decision Tree and Random Forest classifiers using Scikit-learn \cite{scikit-learn}. We evaluate our models' accuracy using a Leave-one-subject-out (LOSO) cross-validation approach.

We perform zero-shot cross-lingual binary classification of aphasic vs control subjects, from English to French and Greek respectively. We first test the performance of our models using oracle transcriptions in all languages. Because we introduce a generic end-to-end pipeline, we normalize the target languages' input features using Scikit-learn's Standard scaler \cite{scikit-learn} that was fitted in the source language, i.e., English. 

Then, we investigate the integration of ASR transcripts under two settings. In the first setting, we train our classifier leveraging the English oracle transcripts, and we evaluate it using the ASR transcripts from the low-resource, target languages. In the second setting, we replace the English oracle training data with ASR transcripts in order to develop a fully end-to-end cross-lingual aphasia detection system. With the last experiment we want to explore whether such a system can be developed without any manually transcribed data. This would allow us to benefit from a much greater amount of data for training and potentially improve overall detection performance. 

\section{Results and Discussion}

\subsection{Automatic Speech Recognition Performance}
\label{asrresults}
We present the performance of the ASR models for each language in Table \ref{table:asr_performance}, as measured by Word Error Rate (WER). As expected, ASR models perform worse in aphasic speech than in healthy speech. We observe that the 
English XLSR-53 model outperforms the other two, while the Greek model outperforms the French one in all categories. 
Adding a language model to the ASR system significantly drops WER in English and French, but exhibits inconsistent results in Greek.
\begin{table}[th]
  \caption{Pre-trained XLSR-53 ASR models performance in English, French and Greek (with and without a Language Model)}
  \label{table:asr_performance}
  \centering
  \begin{tabular}{ l l l l  }
    \toprule
    \multicolumn{1}{c}{\multirow{2}{*}{\textbf{Language}}} &
    \multicolumn{3}{c}{\textbf{WER (\%)}}  \\
    \cline{2-4}
    & \multicolumn{1}{c}{\textbf{Total}} &
    \multicolumn{1}{c}{\textbf{Control}} &
    \multicolumn{1}{c}{\textbf{Aphasia}} \\

    \midrule
    english-no-lm     &52.57   &41.45   &62.74      \\
    english-with-lm   &\textbf{47.14}   &\textbf{37.46}   &\textbf{55.98}      \\
    \hline
        french-no-lm    &71.97   &66.67   &85.71     \\
    french-with-lm  &\textbf{66.88}   &\textbf{62.78}   &\textbf{77.52}      \\
    \hline
    greek-no-lm   & \textbf{65.66}   &\textbf{61.98}   &72.24      \\
    greek-with-lm &67.34   &65.11   &\textbf{71.33}    \\

    \bottomrule
  \end{tabular}
  
\end{table} 
\vspace{-15pt}
\subsection{Monolingual Aphasia Detection - English Only}
We compare the performance of various classification models trained and evaluated solely on English control and aphasic speakers, and we present the results in Table \ref{table:monolingual_results}. Three different scenarios are evaluated, namely using oracle transcripts, ASR results, and ASR with LM results for both training and testing. As shown, the SVM classifier outperforms the XGBoost, Decision Tree and Random Forest classifiers in every setting. We conclude that, given our feature set, we can achieve satisfactory results for aphasia detection in English, even when using ASR transcripts.


\begin{table}[th]
  \caption{Comparison of four classifiers in terms of accuracy (\%) in the English-only experiment, using oracle and ASR transcriptions (with and without a Language Model).}
  \label{table:monolingual_results}
  \centering
  \resizebox{\linewidth}{!}{\begin{tabular}{ l c c c }
    \multicolumn{1}{c}{} &
    \multicolumn{3}{c}{\textbf{Accuracy(\%)}} \\
    \cline{2-4}
    \multicolumn{1}{c}{\textbf{Model}} & 
    \multicolumn{1}{c}{\textbf{Oracle}} & 
    \multicolumn{1}{c}{\textbf{ASR}} & 
    \multicolumn{1}{c}{\textbf{ASR with LM}} \\
    \midrule
    SVM             & \textbf{97.45}     & \textbf{93.90}   & \textbf{93.76}\\
    Decision Tree   & 92.06     & 87.38   &89.22\\
    XGBoost         & 96.31   & 93.62  &93.48\\
    Random Forest & 96.03& 92.34 &92.20\\
    \bottomrule
  \end{tabular}}
  
\end{table} 
\vspace{-15pt}

\begin{table}[th]
  \caption{Comparison of four classifiers in terms of accuracy (\%) in the cross-lingual experiment using oracle transcriptions for all languages.}
  \label{table:cross_lingual_results_oracle}
  \centering
  \resizebox{\linewidth}{!}{\begin{tabular}{ l c c c c}
 \multicolumn{1}{c}{} &
    \multicolumn{4}{c}{\textbf{Accuracy(\%)}} \\
    \cline{2-5}
    \multicolumn{1}{c}{\textbf{Experiment}} & 
    \multicolumn{1}{c}{\textbf{SVM}} & 
        \multicolumn{1}{c}{\textbf{Decision Tree}} &
        \multicolumn{1}{c}{\textbf{XGBoost}} &
    \multicolumn{1}{c}{\textbf{Random Forest}} \\
    \midrule
    English $\rightarrow$ French & 48.15& 81.48& \textbf{100} & \textbf{100}\\
    English $\rightarrow$ Greek   & 52.38 & 59.52& \textbf{97.62}& 85.71\\ 
    \bottomrule
  \end{tabular}}
  
\end{table} 
\vspace{-10pt}
 
 \subsection{Cross-Lingual Aphasia Detection using Oracle Transcripts during Training} 
 \label{section:oracle-training}
 
In Table \ref{table:cross_lingual_results_oracle}, we show the results for the cross-lingual experiment using oracle transcripts for all languages. Since both the French and Greek datasets are balanced we only report accuracy. XGBoost classifier outperforms other classifiers in all settings and thus we chose to exclusively use it in the next experiments.
 We achieve almost perfect results in both Greek and French using oracle transcripts. 
 As shown in Table \ref{table:cross_lingual_results_asr_test}, when using ASR transcripts for the test set, our classifier's accuracy drops by almost 19 \% and  14 \% in French and Greek respectively. The addition of the language model, while increasing accuracy in French, actually decreases performance in Greek.
 
 \begin{table}[th]
  \caption{Zero-shot cross-lingual aphasia detection results from English to French and Greek. The classifier is trained on oracle transcripts and tested on ASR output.}
  \label{table:cross_lingual_results_asr_test}
  \centering
  \resizebox{\linewidth}{!}{\begin{tabular}{ l l l }
    \toprule
    \multicolumn{1}{c}{\textbf{Experiment}} & 
    \multicolumn{1}{c}{\textbf{Transcriptions}} &
    \multicolumn{1}{c}{\textbf{Accuracy (\%)}}  \\
    \midrule
    English $\rightarrow$ French &  Oracle $\rightarrow$ Oracle&100\\
                                 &  Oracle $\rightarrow$ ASR&77.78\\
                                 &  Oracle $\rightarrow$ ASR with LM& 81.48 \\
                                 \hline
    English $\rightarrow$ Greek  &  Oracle $\rightarrow$ Oracle&97.62\\
                                 &  Oracle $\rightarrow$ ASR&83.33\\
                                 &  Oracle $\rightarrow$ ASR with LM&66.67\\
    \bottomrule
  \end{tabular}}
  
\end{table} 
\vspace{-5pt}
 \subsection{Cross-Lingual Aphasia Detection using ASR Transcripts during Training}
In Table \ref{table:cross_lingual_results_asr_all}, we demonstrate that training our classifier on English ASR-derived transcriptions produces results comparable to those obtained using oracle transcriptions (Section \ref{section:oracle-training}), supporting our hypothesis that an end-to-end pipeline is achievable in the cross-lingual setting. Specifically, we achieve up to 88.88 \% and 76.19 \% accuracy in French and Greek respectively, depending on whether we use a language model for these languages. The addition of the language model for French and Greek worsened accuracy results in almost every case. 
Based on our results so far, we are unable to reach a resolution regarding the integration of a language model. This could be because the language models were trained on a generic data corpus (Wikipedia) rather than a dataset-specific domain corpus.
 
 \begin{table}[th]
  \caption{Zero-shot cross-lingual aphasia detection results from English to French and Greek. For each language pair, we evaluate the performance of our classifier in terms of accuracy using ASR-derived transcriptions for training too.}
  \label{table:cross_lingual_results_asr_all}
  \centering
  \resizebox{\linewidth}{!}{\begin{tabular}{ l l c  }
    \toprule
    \multicolumn{1}{c}{\textbf{Experiment}} & 
    \multicolumn{1}{c}{\textbf{Transcriptions}} &
    \multicolumn{1}{c}{\textbf{Accuracy (\%)}}  \\
    \midrule
    English $\rightarrow$ French  &ASR$\rightarrow$ Oracle&81.48\\
                                 &ASR $\rightarrow$ ASR&77.78\\
                                 &ASR $\rightarrow$ ASR with LM&62.96\\
                                 &ASR with LM$\rightarrow$ Oracle&88.88\\
                                 &\textbf{ASR with LM$\rightarrow$ ASR} &\textbf{88.88}\\ 
                                 &ASR with LM$\rightarrow$ ASR with LM&77.77\\
    \hline                             
    English $\rightarrow$ Greek &ASR$\rightarrow$ Oracle&83.33\\
                                 &ASR $\rightarrow$ ASR&73.81\\
                                 &ASR $\rightarrow$ ASR with LM&66.67\\
                                 &ASR with LM$\rightarrow$ Oracle&92.86\\
                                 &\textbf{ASR with LM$\rightarrow$ ASR}&\textbf{76.19}\\ 
                                 &ASR with LM$\rightarrow$ ASR with LM&76.19\\

    \bottomrule
  \end{tabular}}
\end{table} 
\vspace{-5pt}
\section{Conclusion and Future Work}

In this work, we established an end-to-end pipeline for cross-lingual aphasia detection by using language-agnostic linguistic features from automatic speech recognition transcripts. 
We achieved up to 81 \% and 83 \% detection accuracy in French and Greek respectively, when training our model using human-annotated speech transcripts, and up to 88 \% and 76 \% for the end-to-end case when only using ASR output.
Our results indicate that it is possible to apply transfer learning between two closely related languages even without oracle transcripts in both source and target languages. In the future, we aim to extend our research by developing a similar procedure for aphasia severity assessment and aphasia type classification, as well as study the effects of adapting the ASR and language models to domain-specific data.

\bibliographystyle{IEEEtran}

\bibliography{PlanV_biblio}


\end{document}